\pdfoutput=1

\documentclass[11pt]{article}

\usepackage[final]{acl}

\usepackage{times}
\usepackage{latexsym}
\usepackage{booktabs}
\usepackage{multirow}
\usepackage{subcaption}
\usepackage{tcolorbox}
\usepackage{multirow, colortbl, caption}
\usepackage[T1]{fontenc}

\usepackage[utf8]{inputenc}

\usepackage{microtype}

\usepackage{inconsolata}

\usepackage{graphicx}
\usepackage{comment}

 \title{Deploying Tiny LVLM Judges for Real-World Evaluation of Chart Models: Lessons Learned and Best Practices
 \vspace{-50pt}}
 \author{Md Tahmid Rahman Laskar\textsuperscript{\textdaggerdbl,}\thanks{\hspace{0.115cm} Contact Emails: \{tahmedge,enamulh,jhuang\}@yorku.ca},  \textbf{Mohammed Saidul Islam\textsuperscript{\textdaggerdbl}}\textbf{,}   \textbf{Ridwan Mahbub\textsuperscript{\textdaggerdbl}}\textbf{,} \\   \textbf{Mizanur Rahman}\textsuperscript{\textdaggerdbl}\textbf{,}   
\textbf{Md Amran Hossen Bhuiyan\textsuperscript{\textdaggerdbl}}\textbf{,} 
  \textbf{Israt Jahan\textsuperscript{\textdaggerdbl,}}\textbf{, } \\
 \textbf{Mir Tafseer Nayeem\textsuperscript{\textsection}}\textbf{,}  
 \textbf{Shafiq Joty\textsuperscript{\textdollar}}\textbf{, }
 \textbf{Enamul Hoque\textsuperscript{\textdaggerdbl,}}\footnotemark[1]\textbf{, }
 \textbf{Jimmy Xiangji Huang\textsuperscript{\textdaggerdbl,}\footnotemark[1]} \\
            {\textsuperscript{\textdaggerdbl}York University, 
                     \textsuperscript{\textsection}University of Alberta, 
\textsuperscript{\textdollar}Salesforce AI Research}
          }

\begin{document}
\maketitle
\begin{abstract}
Large Vision-Language Models (LVLMs) with only 7B parameters have shown promise as automated judges in chart comprehension tasks. However, tiny models ($\leq$2B parameters) still perform poorly as judges, limiting their real-world use in resource-constrained settings. To address this, we propose two approaches to ensure cost‑efficiency in real-world evaluation of chart models: (i) multi‑criteria prompting, which combines separate evaluation criteria into a single instruction, and (ii) domain‑adaptive transfer learning, in which we address the low-resource problem by fine‑tuning a 2B‑parameter LVLM on synthetic judgments in a chart dataset to create the \textbf{ChartJudge-2B} model. Experiments show that multi-criteria prompting exposes robustness gaps, which led to a huge drop in performance for 7B models, including specialized LVLM judges like LLaVA‑Critic. In addition, we find that by fine-tuning a tiny LVLM (i.e., ChartJudge-2B), we can effectively transfer knowledge from one dataset to another to make it a more specialized model. Our fine-grained analysis across chart types and query complexities offers actionable insights into trade-offs between model size, prompt design, and transferability—enabling scalable, low-cost evaluation for chart reasoning tasks. 
\end{abstract}
\section{Introduction}
Charts are widely used to communicate data, but understanding them requires multimodal reasoning that integrates visual and textual cues~\cite{hoque2024natural}. 
While LVLMs show strong performance on such tasks across automatic metrics, real-world deployment often requires qualitative evaluation, which typically relies on costly and time-consuming human judgment~\cite{islam2024large}.

\begin{figure}[t!]
    \centering
    \includegraphics[width=\linewidth]{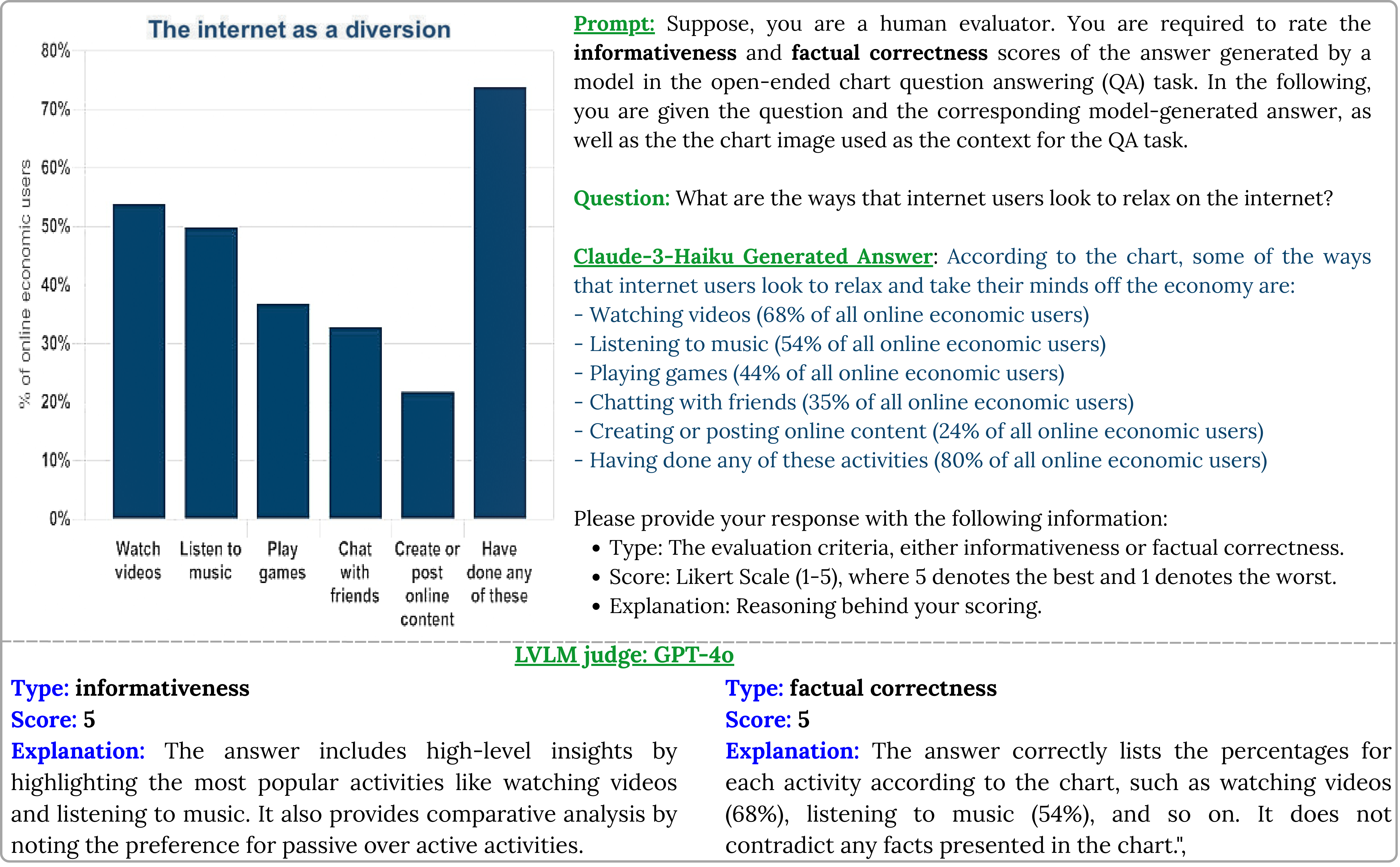}
    \caption{GPT-4o as the judge evaluating \textit{Claude-3-Haiku} model answer in the OpenCQA dataset across multiple criteria (informativeness, factual correctness).
    }
    \label{fig:intro_figure}
\end{figure}

Recently, \citet{laskar2025judgingjudgeslargevisionlanguage} proposed the utilization of LVLMs as the judge to evaluate other LVLMs in downstream chart tasks (see Figure \ref{fig:intro_figure} for an example of an LVLM judge for charts), with certain 7B LVLMs achieving judgment performance closer to powerful LVLMs like GPT-4o \cite{openai2023gpt4} or LLaVA-Critic-70B \cite{xiong2024llavacritic}. This creates opportunities to use open-source LVLM judges in real-world scenarios where industries do not share sensitive proprietary charts with closed-source LLM providers.  However, tiny LVLMs ($\leq$2B parameters) still underperform, limiting their applicability in resource-constrained settings, since available tools for optimizing LVLMs for inference are much less available than text-only LLMs~\cite{shinde2025survey}. 

\begin{figure*}
    \centering
    \includegraphics[width=\linewidth]{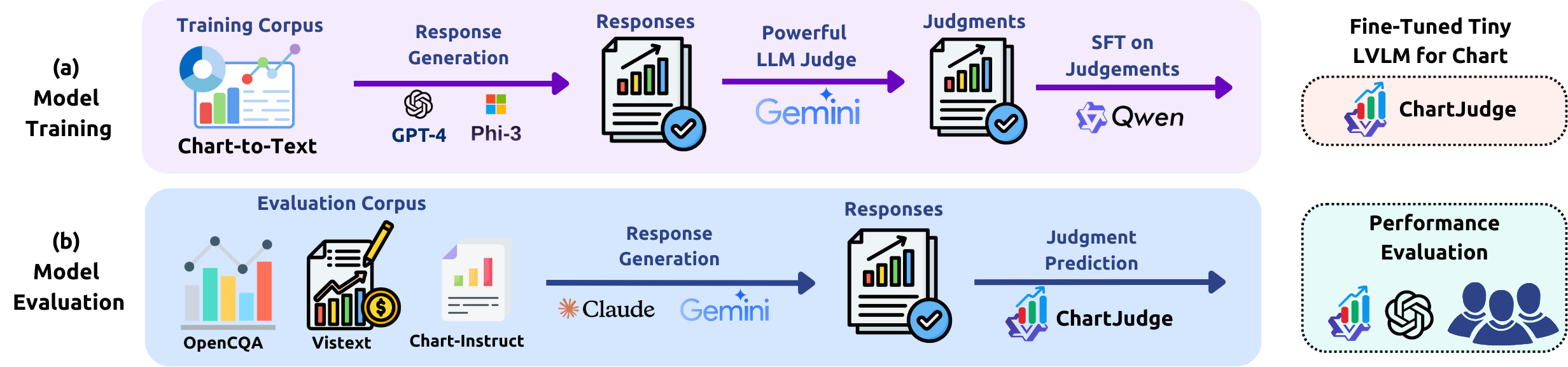}
\caption{Our Fine-Tuning Approach. (a) At first, responses generated by various LVLMs (e.g., GPT-4, Phi-3) for chart captioning  (e.g., Chart-to-Text dataset) are judged on diverse criteria by a different LVLM (e.g., Gemini-1.5-Pro). A small LVLM (e.g., Qwen2-VL-2B-Instruct) is then fine-tuned on these judgments to create \textbf{ChartJudge-2B}. (b) For evaluation, responses from LVLMs (e.g., Claude, Gemini) on chart benchmarks are judged by \textbf{ChartJudge-2B} and compared with a larger LVLM (e.g., GPT-4o) or human ratings.}
    \label{fig:ft_figure}
\end{figure*}

To address these challenges, we explore two strategies for cost-efficient LVLM-based evaluation considering real-world industrial settings: \textbf{(i)} \textit{multi-criteria prompting}, which fuses multiple evaluation dimensions into a single query to reduce inference latency \& cost, and \textbf{(ii)} \textit{domain-adaptive transfer learning}, where a 2B-parameter LVLM is fine-tuned on synthetic judgments from a stronger model. The result is \textbf{ChartJudge-2B}, a lightweight evaluator tailored for chart tasks (Figure~\ref{fig:ft_figure}). Our experiments show that multi-criteria prompting exposes significant performance drops in existing LVLMs—not only in 7B models like the specialized LLaVA-Critic-7B, but also in its larger variant, LLaVA-Critic-70B~\cite{xiong2024llavacritic}. In contrast, ChartJudge-2B, with only 2B parameters, generalizes effectively across chart datasets, offering a cost-effective alternative for 
evaluation. Our work makes \emph{three} key contributions: 

\begin{itemize}

\item  We demonstrate that ChartJudge-2B can effectively transfer knowledge across chart datasets, even when distilled from a different LVLM judge than the one used for evaluation, making it a robust, cost-efficient evaluator in low-resource settings.

\item We show that multi-criteria prompting reveals significant robustness issues and performance drops in open-source LVLMs, including specialized judges like LLaVA-Critic. However, a tiny LVLM can achieve strong performance in multi-criteria prompting when fine-tuned even on limited amount of data. 

\item We provide practical insights on model selection, prompt design, model errors, and transfer strategies, helping guide scalable deployment of LVLM judges for real-world chart reasoning tasks.

\end{itemize}

As a secondary contribution, our code and the data will be made publicly available here: \url{https://github.com/tahmedge/chart_lvlm_judge}. 

\section{Related Work}

The complexity of chart understanding tasks like question answering \cite{hoque2022chart, masry2022chartqa, masry-etal-2025-chartqapro,open-CQA} and captioning \cite{chart-to-text-acl,2023-vistext} has led to the creation of various specialized chart-specific models \cite{han2023chartllama,masry2025chartgemma}. However, evaluating these specialized models is a significant hurdle due to the dependence on humans. To overcome this, recent research explores using LVLMs as judges for the chart domain \cite{laskar2025judgingjudgeslargevisionlanguage}. 

Although early LVLM judges like Prometheus-VL \cite{lee2024prometheusvl} found text-rich visuals like charts challenging for evaluation, \citet{laskar2025judgingjudgeslargevisionlanguage} demonstrate that many recently proposed open-source LVLMs with only 7B parameters can achieve performance comparable to GPT-4o. This makes it possible to even use smaller open-source LVLMs in real-world scenarios where industries restrict the use of closed-source API models for evaluation on proprietary charts.  However, tiny LVLMs (under 2B) still lag behind, limiting their viability in resource-constrained environments.

To address this gap, in this paper, we investigate various strategies to minimize the inference cost of using open-source LVLMs in real-world scenarios. Due to the effectiveness of techniques like prompt engineering \cite{islam2024large} and fine-tuning of smaller models \cite{masry2025chartgemma} that help achieve performance gain while reducing inference cost, 
we seek to find the best strategy to utilize these techniques 
in the context of the LVLM judge. 
\section{Methodology}
\subsection{Proposed Approach}
To achieve cost-effective inference in real-world industrial settings, we proposed two approaches.
\paragraph{(i) Fine-Tuning Approach:} The lack of human-annotated judgment dataset in the chart domain prohibits the training of any LVLMs-as-the-judge for chart-related tasks. Since early research shows that annotations generated by powerful closed-source LVLMs correlate well with humans~\cite{lee2024prometheusvl,chen2024mllm}, we propose a \textit{domain-adaptive fine-tuning approach} \cite{laskar2020wsl,laskar2022domain,laskar-etal-2025-improving} that leverages knowledge distillation by first generating judgments in a given chart dataset on different LVLM-generated responses using a larger LVLM. Then, the smaller LVLM is fine-tuned on this judgment data for better adaptation in the chart domain (see Figure \ref{fig:ft_figure}). 

\paragraph{(ii) Prompting-Based Approach:}
Since LVLM fine-tuning requires good computing resources, it is often difficult to fine-tune LVLMs in resource-constrained scenarios \cite{shinde2025survey}. In such cases, prompt engineering can help improve the performance of LVLM without 
any additional training.  
However, evaluating with separate prompts for different evaluation criteria increases both inference latency and cost \cite{laskar-etal-2023-systematic,laskar-etal-2024-query}. In the {prompting-based approach}, our main objective is to reduce evaluation cost by minimizing the overall inference tokens and  GPU time. To this end, we propose the \textit{multi-criteria prompting} approach that combines multiple evaluation criteria within a single prompt (see Appendix \ref{appendix_prompt} for a sample prompt) for the LVLM judges to generate judgments for all criteria in their response (see Appendix \ref{appendix_response} for a sample response).
\subsection{Evaluation Protocol}
Given a chart and model-generated response(s), we follow the work of \citet{laskar2025judgingjudgeslargevisionlanguage} to construct prompts, 
evaluation rubric, and evaluation metrics.

\paragraph{Evaluation Rubric and Metrics:} 
\label{eval_rubric_metrics}
We construct prompts for both \textit{\textbf{pairwise}} and \textit{\textbf{pointwise}} evaluation. The constructed prompts contain samples \textit{\textbf{with reference}} as well as 
\textit{\textbf{without reference}}, for diverse evaluation criteria: \textit{\textbf{factual correctness}}, \textit{\textbf{informativeness}}, \textit{\textbf{relevance}}, and \textit{\textbf{multidimensional criteria}} (e.g., overall response quality). In line with the prior work, we also use the following evaluation metrics: \textit{\textbf{Judgment Accuracy}}, \textit{\textbf{Error Distance}}, \textit{\textbf{Bias Evaluation}}, \textit{\textbf{Instruction Following Evaluation Capability}}, and \textit{\textbf{Format Adherence}}. We refer readers to the Appendix \ref{definition_metrics} for the definitions of these metrics. 
\paragraph{Evaluation Datasets:} For evaluation, we use the following three datasets:
 (i) \textbf{OpenCQA} \cite{chart-to-text-acl} for open-ended chart QA.

(ii) \textbf{VisText} \cite{2023-vistext} for chart captioning consisting of L1 (describing chart's structural elements) and L2/L3 captions (providing key insights in the data). 

 (iii) \textbf{Chart-Instruct-Eval} \cite{laskar2025judgingjudgeslargevisionlanguage} that assesses LVLM judges on evaluating the instruction following capabilities of other LVLMs in chart-related tasks. 

The judgment annotations in these datasets are collected from \cite{laskar2025judgingjudgeslargevisionlanguage}, with OpenCQA and VisText containing 100K judgment data, and Chart-Instruct-Eval containing 400 samples. More specifically, for OpenCQA and VisText, we collect the outputs generated by \citet{islam2024large} using \textit{Gemini-1.0-Pro}  \cite{geminiteam2023gemini} and \textit{Claude-3-Haiku} \cite{Claude}, and use the judgment scores annotated by \cite{laskar2025judgingjudgeslargevisionlanguage} using \textit{GPT-4o} \cite{openai2023gpt4} and \textit{LLaVA-Critic-70B} \cite{xiong2024llavacritic} models for diverse evaluation rubrics across 100K judgment data. For Chart-Instruct-Eval, we use the existing annotation in the dataset. For the multi-criteria evaluation, we use the OpenCQA dataset. 

\definecolor{open_models_below_4B}{RGB}{185, 235, 255}
\definecolor{open_models_7B_12B}{RGB}{255, 219, 187}
\definecolor{models}{RGB}{243, 243, 243}

\definecolor{open_models_below_4B}{RGB}{185, 235, 255}
\definecolor{open_models_7B_12B}{RGB}{255, 219, 187}
\definecolor{data_red}{RGB}{255, 204, 204}
\definecolor{data_green}{RGB}{229, 255, 229}
\definecolor{data_blue}{RGB}{230, 230, 255}
\definecolor{data_deepblue}{RGB}{224, 239, 255}
\begin{table}[t!]
\centering
\scriptsize
\begin{tabular}{lcc}
\toprule
\textbf{Type} & \textbf{\# Single-Criterion} & \textbf{\# Multi-Criteria} \\ 
\midrule

\textbf{\textit{Source-wise}} & & \\ 
\rowcolor{models} Statista & 6,898 & -- \\ 
\rowcolor{models} Pew      & 2,827 & 2,827 \\ 
\midrule
{\textbf{\textit{Pointwise Labels}}} & & \\ 
\rowcolor{open_models_below_4B!50} 1 & 801  & 510  \\ 
\rowcolor{open_models_below_4B!50} 2 & 1,000 & 548 \\ 
\rowcolor{open_models_below_4B!50} 3 & 1,000 & 414 \\ 
\rowcolor{open_models_below_4B!50} 4 & 1,000 & 179 \\ 
\rowcolor{open_models_below_4B!50} 5 & 1,000 & 113 \\ 
\midrule
{\textbf{\textit{Pairwise Labels}}} & &  \\ 
\rowcolor{open_models_7B_12B!50} Tie    & 2,000 & 268 \\ 
\rowcolor{open_models_7B_12B!50} Model A & 1,500 & 568 \\ 
\rowcolor{open_models_7B_12B!50} Model B & 1,424 & 226 \\ 
\bottomrule
\end{tabular}
\caption{Fine-tuning dataset statistics for single-criterion and multi-criteria settings. Here, `\#' denotes total number of samples.}
\label{tab:dataset_statistics}
\end{table}

\subsection{ChartJudge-2B}

We fine-tune the tiny LVLM \texttt{Qwen2-VL-2B-Instruct\footnote{The Qwen2-VL-2B-Instruct model was selected due to its strong performance as a 2B parameter model across diverse tasks~\cite{wang2024qwen2}. }}~\cite{wang2024qwen2} to build the \textbf{ChartJudge-2B} model. We describe our fine-tuning data construction procedure below. 

We constructed a training dataset comprising both pointwise and pairwise evaluations of chart image summaries sourced from the \emph{Chart-to-Text} benchmark \cite{chart-to-text-acl}. However, the {Chart-to-Text} dataset is a chart captioning benchmark, whereas our evaluation datasets contain both question answering (QA) and chart captioning. Since {Chart-to-Text} contains data from Statista and Pew, we generate synthetic questions on its Pew split to use it as an open-ended QA task,  while using the Statista split for chart captioning without any modification. 
For question generation, we prompt the \textbf{Gemini-1.5-Pro} \cite{geminiteam2024gemini15unlockingmultimodal} model to generate a question by providing the chart and its corresponding gold reference as input (see Appendix \ref{appendix_pew_question} for the prompt). 

To mitigate potential \textit{teacher model} bias, 
we intentionally decoupled the models used at each stage of our pipeline. For the fine-tuning data, responses were generated by \textit{GPT-4V} \cite{openai2023gpt4} and \textit{Phi-3}\cite{abdin2024phi}. The judgments on these responses, which serve as training labels, were then created by a different model, \textit{Gemini-1.5-Pro} \cite{geminiteam2024gemini15unlockingmultimodal}. For the final evaluation of our \textit{ChartJudge-2B} model, we again used a distinct set of models: responses were generated by \textit{Claude-3-Haiku}~\cite{Claude} and \textit{Gemini-1.0-Pro}, and the reference judgments were provided by \textit{GPT-4o} and \textit{LLaVA-Critic-70B}. This is done to ensure robustness across different LVLM-generated responses and judgments by using different models for training and evaluation. 


\begin{table}[t]
    \centering
  \setlength{\tabcolsep}{2.7pt}
    {
        \resizebox{7.7cm}{!}{
        \begin{tabular}{l|cc|cc|cc}
            \toprule
         
            \multicolumn{1}{c}{\multirow{2}{*}{\textbf{Model}}} &
              \multicolumn{2}{c}{\textbf{OpenCQA}} &
              \multicolumn{2}{c}{\textbf{VisText L1}} &
              \multicolumn{2}{c}{\textbf{VisText L2/L3}} \\
              \cmidrule{2-7}

            \multicolumn{1}{c}{} &
              \textbf{Pair ($\uparrow$)} & \textbf{Point ($\downarrow$)} & \textbf{Pair ($\uparrow$)} & \textbf{Point ($\downarrow$)} &
              \textbf{Pair ($\uparrow$)} & \textbf{Point ($\downarrow$)} 
            \\
            \midrule
       \rowcolor{models} ChartJudge-2B &  61.7 & 1.3 & 64.6 & 1.6 & 52.3 & 1.3 \\
\rowcolor{open_models_below_4B!50} Qwen2‑VL‑2B‑Instruct
          &   54.0  &   1.0
        &    27.2   
     &    2.1
       &   3.0

    &   0.9\\
\rowcolor{open_models_below_4B!50} Phi‑3.5‑Vision‑3.8B‑Instruct
            & 50.7 &  0.8
            & 69.5  &  1.5
            & 49.5

            &  1.0\\
\rowcolor{open_models_below_4B!50} XGen‑MM‑Phi3‑3.8B‑Instruct
            & 71.6    &  0.9
            & 75.4      &  1.4
            & 70.7   &  0.7

         \\

\rowcolor{open_models_7B_12B!50} Qwen2‑VL‑7B‑Instruct
            & 66.9     &  0.7
            & 57.6     &  0.6
            & 70.0    &  0.7

        \\
\rowcolor{open_models_7B_12B!50} InternLM‑Xcomposer2d5‑7B
            & 64.5     & 0.9
           & 72.0    &  0.9
           & 75.6 &  0.7

           \\
\rowcolor{open_models_7B_12B!50} LLaVA‑Next‑v1.6‑Mistral‑7B
            & 75.9       &  0.8
            & 75.1   &  1.4
            & 75.1    & 0.9

        \\

      
   %
         

            \rowcolor{open_models_7B_12B!50} {LLaVA‑Critic‑7B}
            & \textbf{79.5}    & \textbf{0.5}
            & \textbf{79.1}  
            & \textbf{0.5}
            & \textbf{77.1}
            & \textbf{0.6}\\
            \bottomrule
        \end{tabular}
        }
    }
    \caption{\small{Average results (judgment accuracy for pairwise, error distance for pointwise) of different models (\colorbox{open_models_below_4B!50}{below 4B}, \colorbox{open_models_7B_12B!50}{above 4B}) from \citet{laskar2025judgingjudgeslargevisionlanguage} compared to our   
   \colorbox{models}{ChartJudge-2B} model.`$\uparrow$' = higher better; `$\downarrow$' = lower  better. 
  }
  }
    \label{tab:overall_avg_ft_results}
\end{table}

\begin{table}[t]
\scriptsize
\setlength{\tabcolsep}{1.5pt} 
    \centering
    \resizebox{7.75cm}{!}{
    \begin{tabular}{l|c|c|c}
        \toprule
        \textbf{Model} & \textbf{Instruction ($\uparrow$)} & \textbf{Format ($\uparrow$)} & \textbf{Bias ($\downarrow$)} \\
        \midrule
        \rowcolor{models} ChartJudge-2B & 36.5 & 95.9 & 79.5 \\
        \rowcolor{open_models_below_4B!50} Qwen2-VL-2B-Instruct & 13.5 & 78.9 & 63.5 \\
        \rowcolor{open_models_below_4B!50} Phi-3.5-Vision-3.8B-Instruct & 49.0 & 83.3 & 64.7 \\
        \rowcolor{open_models_below_4B!50} XGen-MM-Phi3-3.8B-Instruct & 72.5 & 97.6 & 71.8 \\
        \rowcolor{open_models_7B_12B!50} Qwen2-VL-7B-Instruct & \textbf{87.0} & 98.6 & \textbf{28.7} \\
    \rowcolor{open_models_7B_12B!50} InternLM‑Xcomposer2d5‑7B
            & 24.5     & 35.9
           & 30.2 \\
        \rowcolor{open_models_7B_12B!50} LLaVA-Next-v1.6-Mistral-7B & 27.0 & 98.9 & 74.4 \\
        \rowcolor{open_models_7B_12B!50} LLaVA-Critic-7B & 45.5 & \textbf{99.7} & 58.0 \\
  
        \bottomrule
    \end{tabular}
    }
    \caption{Performance of our fine-tuned \colorbox{models}{ChartJudge-2B} and other baselines for Instruction Following evaluation capability on Chart-Instruct-Eval, Format Adherence (average across all datasets), and Bias (average of Length and Position Bias across all datasets). `$\uparrow$' = higher is better; `$\downarrow$' = lower is better.}
    \label{tab:format_instruction_accuracy_comparison}
\end{table}

In the pointwise setting, each summary was assigned a rating ranging from 1 to 5, accompanied by a textual justification for the rating. In the pairwise setting, the model was prompted to compare two responses (Model A and Model B) and select the better one, or indicate a tie if no clear preference could be determined, along with an explanatory rationale for the decision. We present our fine-tuning dataset statistics in Table \ref{tab:dataset_statistics} (more information on the dataset is also provided in Appendix \ref{chart_type_statistics}). Below, we describe the dataset based on the \textit{single-criterion} and the \textit{multi-criteria} settings. 
 
\textbf{Single-Criterion:} 
For the {single-criterion} scenario, our fine-tuning dataset consists of 9,725 samples. While the instances in the dataset are randomly sampled, we avoid any bias towards a particular type of evaluation type/criteria by defining a data distribution schema such that the dataset does not contain too many samples of a particular evaluation type/criteria. 

\textbf{Multi-Criteria:} 
We construct a dataset of 2,827 samples by utilizing \emph{informativeness} and \emph{factual correctness} as the evaluation criteria in the multi-criteria prompt. 
This dataset also comprises both pointwise and pairwise evaluations of chart images and sourced from the Pew split of our single-criterion evaluation dataset. Moreover, similar to the single-criterion scenario, 
the annotations were generated using the \textit{Gemini-1.5-Pro} model. 

\section{Experiments}
\subsection{Experimental Settings}
\label{hyperparameters}
For inference, we set the temperature to 1.0 to ensure diversity in LLM responses. All other decoding parameters were set to their default values, as provided by the respective API providers, OpenAI, Google Vertex AI, and Anthropic for closed-source models, and HuggingFace \cite{wolf2019huggingface} for open-source models. Given that the task did not require generating lengthy outputs, we limited the maximum output length to 300 tokens. For fine-tuning, we trained the Qwen2-VL-2B-Instruct for three epochs using a batch size of 2, with the learning being tuned between $1e^{-4}$ to $2e^{-5}$. All experiments were run using two NVIDIA H100 GPUs.

\begin{table*}[t!]
  \centering
        \resizebox{16cm}{!}
        {
    \setlength{\tabcolsep}{4pt}
        \begin{tabular}{l|ccc|ccc|c|ccc|ccc|c}
            \toprule
            \multicolumn{1}{c}{} &
              \multicolumn{7}{c}{\textbf{Single-Criterion}} &
              \multicolumn{7}{c}{\textbf{Multi-Criteria}}\\
            \cmidrule(lr){2-8}\cmidrule(lr){9-15}
            \multicolumn{1}{c}{\multirow{2}{*}{\textbf{Model}}} &
              \multicolumn{3}{c}{\textbf{Pairwise ($\uparrow$)}} &
              \multicolumn{3}{c}{\textbf{Pointwise} ($\downarrow$)} &
              \multicolumn{1}{c}{\textbf{Format Following ($\uparrow$)}} &
              \multicolumn{3}{c}{\textbf{Pairwise} ($\uparrow$)} &
              \multicolumn{3}{c}{\textbf{Pointwise} ($\downarrow$)} &
              \multicolumn{1}{c}{\textbf{Format Following} ($\uparrow$)}\\
            \cmidrule(lr){2-4}\cmidrule(lr){5-7}\cmidrule(lr){8-8}\cmidrule(lr){9-11}\cmidrule(lr){12-14}\cmidrule(lr){15-15}
            \multicolumn{1}{c}{} &
              \textbf{FC} & \textbf{I} & \textbf{Avg.} &
              \textbf{FC} & \textbf{I} & \textbf{Avg.} &
              \textbf{Overall Avg.} &
              \textbf{FC} & \textbf{I} & \textbf{Avg.} &
              \textbf{FC} & \textbf{I} & \textbf{Avg.} &
              \textbf{Overall Avg.} \\
            \midrule
\rowcolor{models} ChartJudge-2B (Multi-Criteria) & 57.70 & 74.99 & 66.35 & 2.15 & 2.00 & 2.08 & 95.83 & \textbf{30.87} & 62.85 & \textbf{46.86} & 1.13 & 2.08 & 1.61 & 86.60
 \\
\rowcolor{open_models_below_4B!50} Qwen2-VL-2B-Instruct & 29.14 & 69.93 & 49.54 & 1.19 & 0.95 & 1.07 & 81.10 & 5.15 & 39.17 & 22.16 & 3.61 & 1.89 & 2.75 & 44.29 \\
\rowcolor{open_models_below_4B!50} Phi-3.5-vision-3.8B-instruct &  29.66 & 64.34 & 47.00 & 0.80 & 0.83 & 0.82 & 79.99 & 21.65 & 71.83 & 46.74 & \textbf{1.03} & \textbf{0.99} & \textbf{1.01} & \textbf{92.28} \\
\rowcolor{open_models_below_4B!50} XGen-MM-Phi3-3.8B-Instruct &  57.13 & 76.63 & 66.88 & 1.09 & 0.99 & 1.04 & 96.33  &  19.80 & 68.33 & 44.07 & 1.46 & 1.07 & 1.27 & 85.54 \\
\rowcolor{open_models_7B_12B!50}Qwen2-VL-7B-Instruct &  60.11 & 73.37 & 66.74 & 0.89 & 0.88 & 0.89 & 97.78  &  10.36 & \textbf{70.88} & 40.62 & 1.18 & 1.17 & 1.18 & 74.98 \\
\rowcolor{open_models_7B_12B!50}InternLM-Xcomposer2d5-7B &  60.38 & 68.80 & 64.59 & 0.74 & 0.73 & 0.74 & 94.01 &  7.78 & 27.48 & 17.63 & 3.41 & 3.54 & 3.48 & 34.88 \\
\rowcolor{open_models_7B_12B!50}LLaVA-Next-v1.6-Mistral-7B &  62.85 & 81.41 & 72.13 & 0.98 & 0.85 & 0.92 & 98.57 &  0.51 & 1.34 & 0.93 & 4.86 & 4.88 & 4.87 & 2.29 \\
\rowcolor{open_models_7B_12B!50}LLaVA-Critic-7B &  \textbf{69.65} & \textbf{83.90} & \textbf{76.78} & \textbf{0.70} & \textbf{0.58} & \textbf{0.64} & \textbf{99.57} &  0.00 & 0.00 & 0.00 & 5.00 & 5.00 & 5.00 & 0.00 \\
            \bottomrule
        \end{tabular}
        }
    \caption{Results in OpenCQA for Single and Multi-Criteria (judgment accuracy for pairwise, error distance for pointwise). 
   Here, `FC' denotes `Factual Correctness' and `I' denotes `Informativeness'. `$\uparrow$' = higher is better; `$\downarrow$' = lower is better.}  \label{tab:overall_avg_prompt_results}
\end{table*}

\subsection{Results and Discussion}
We design a parsing script \cite{laskar-etal-2024-systematic} to extract the target outputs from the LVLM-judge-generated judgments. Below, we demonstrate our findings. 
\subsubsection{Fine-Tuning Results}  As shown in Table~\ref{tab:overall_avg_ft_results}, our fine-tuned model, \textbf{ChartJudge-2B}, significantly improves the performance of Qwen2-VL-2B-Instruct in pairwise evaluation across all datasets. It also outperforms larger models in several settings, including Qwen-7B on Vistext L1, and Phi-3.8B on OpenCQA and Vistext L2/L3. 

However, it does not necessarily improve the performance in pointwise scenarios (except in Vistext L1). One possible reason for this discrepancy could be because of using different LVLM judgment scores as the reference during training (Gemini-1.5-Pro) and evaluation (GPT-4o/LLaVA-Critic-70B), leading to the alignment variation on the judgment scores. Despite these variations in absolute scores, we found a positive Spearman's rank correlation \cite{schober2018correlation} between ChartJudge-2B and larger models in how they score examples in pointwise evaluation. This indicates a similar alignment in scoring behavior and relative ranking, even if the exact scores differ.


Moreover, we find in Table \ref{tab:format_instruction_accuracy_comparison} that the instruction following evaluation capability and format following capability are also increased by a large margin for the fine-tuned Qwen2-VL-2B-Instruct model, 
although the overall bias also increased. This could be a limitation of fine-tuned LVLM judges, 
since the LLaVA-Critic-7B model, which achieves the best result in Table \ref{tab:overall_avg_ft_results}, also demonstrates high bias. 

\begin{figure}[t]
    \centering
    \includegraphics[scale = 0.161]{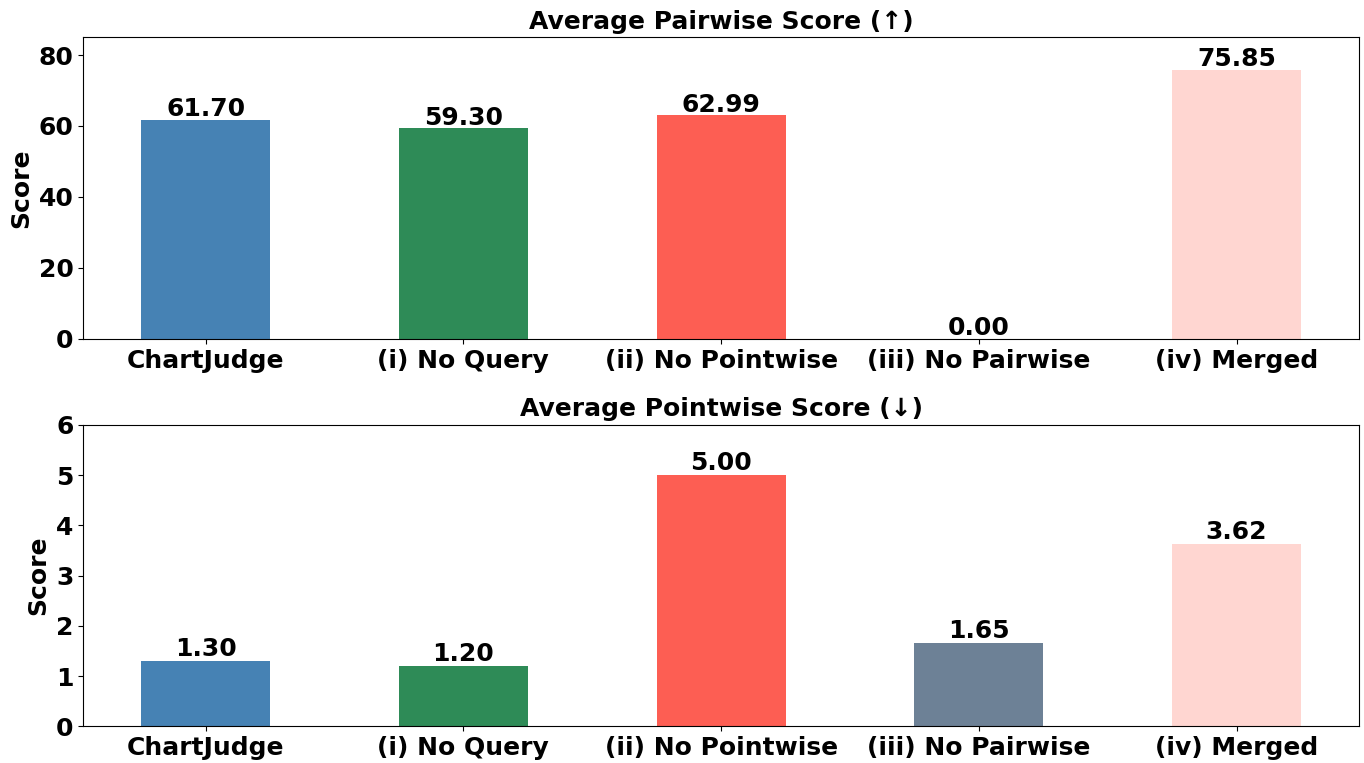}
    \caption{Ablation results for the ChartJudge-2B model in OpenCQA via training  (i) without query, (ii) only pointwise samples, (iii) only pairwise samples, and (iv) merged version of only pairwise and pointwise models.}
    \label{fig:ablation}
\end{figure}

\subsubsection{Performance Comparison: Multi-Criteria vs Single-Criterion} 
We construct multi-criteria prompts in the Pew split of Chart-to-Text and fine-tune the Qwen2-VL-2B-Instruct model only on this data. 
From the results presented in Table \ref{tab:overall_avg_prompt_results}, we find that most existing LVLMs perform quite poorly in multi-criteria settings. However, our fine-tuned model outperforms most other baselines in this setting, especially in pairwise evaluation. This demonstrates that by fine-tuning on limited multi-criteria judgment data, we can achieve better adaptation using a tiny LVLM, offering further optimization of the inference cost. Surprisingly, the best-performing LLaVA-Critic-7B model achieves the worst result in the multi-query setting. Our analysis shows that LLaVA-based models, LLaVA-Next, and LLaVA-Critic (both 7B \& 70B), fail to follow the multi-criteria instruction (see Table \ref{tab:error_examples}), leading to poor results. This may happen due to the training data of LLaVA-based models missing multi-criteria judgments.
\subsubsection{Ablation Test Results} We conduct ablation tests using 
ChartJudge-2B 
to study the following:

\textbf{(a)} What is the effect of removing synthetic queries in our fine-tuning dataset?

\textbf{(b)} If we train the model using only pointwise or pairwise data, can model ensure robustness when evaluated on the data in which they were not trained?

\textbf{(c)} Is it useful to train multiple models, one for pointwise and another for pairwise, and then apply Model Merging \cite{yang2024model} on each model?  

For model merging, the judge LVLM is created using 
two individual Qwen2-VL-2B-Instruct \cite{bai2023qwenvl} models, one fine-tuned for pair-based judgement and the other fine-tuned for point-based judgement. The implementation was done using the \emph{mergekit}~\cite{goddard-etal-2024-arcees} library, and the linear model merging method was used \cite{wortsman2022model}, considering its simplicity. An additional \texttt{visual.merger.mlp} layer with all weights initialized to $0$ was inserted into both base models to ensure compatibility with the \texttt{mergekit} implementation. The \texttt{weight} parameter, representing the contribution of each model to the averaging process, was set to $1.0$ for both models.

From our ablation test results presented in Figure~\ref{fig:ablation}), we observe the following: 

\textbf{(a)} Removing query relevance drops model performance in the pairwise setting. 

\textbf{(b)} The model loses its judgment capability in different settings when trained only on one type of data (e.g., model trained only on pairwise cannot do judgments in pointwise and vice versa). 

\textbf{(c)} Model merging  helps the models trained on individual settings (only pairwise or pointwise) to boost their performance (for instance, the best result in pairwise is achieved using a merged model). 
\subsubsection{Performance on Human-Annotated Complex Charts and Queries}
Following the work of \citet{mukhopadhyay2024unraveling}, we annotate complex queries and charts in OpenCQA (see Appendix \ref{appendix_complex_human} for annotation details) by two human experts in data science and computer vision to investigate the performance of our fine-tuned model in complex scenarios. Both annotators label 896 samples as having complex charts and queries, where the ChartJudge-2B model achieves 63.3\% judgment accuracy, with its zero-shot 
counterpart, the Qwen2-VL-2B-Instruct model achieving 54.5\% accuracy.

\subsubsection{Robustness and Generalization}
\paragraph{Robustness on Weak Base Models:}
To test the robustness of our domain adaptation approach, we applied it to a base model known to be a very weak LVLM judge. We selected PaliGemma-3B, which achieves 0\% accuracy as a zero-shot chart judge in the work of \cite{laskar2025judgingjudgeslargevisionlanguage}. After fine-tuning the model with our proposed method, its pairwise accuracy on the Vistext dataset dramatically improved from 0\% to 55.9\%. Moreover, the format adherence accuracy is also increased to 77.0\% from 0\%. This experiment demonstrates that our fine-tuning technique can overcome significant inherent weaknesses in diverse base models and that the performance gains are attributable to our domain adaptation strategy.

\paragraph{Out-of-Domain Generalization Performance:}

In our prior experiments, we observe the effectiveness of domain adaptive transfer learning using a 2B parameter model by fine-tuning on a dataset different than our evaluation dataset. Since fine-tuning may also lead to catastrophic forgetting \cite{luo2023empirical}, in this section, we investigate the performance on a completely different image understanding tasks. For our evaluation, we use the ChEBI-20-MM \cite{liu2024scientificmoleculechebi} dataset for \textit{caption generation from molecular images} and evaluate the responses generated by \citet{jahan2025evaluating} using the Claude-3-Haiku model and the Gemini-1.5-Flash model. We then evaluate the responses based on pair-wise comparison to select the better model by considering the relevancy, conciseness, informativeness, and factual correctness of the response (see Appendix \ref{appendix_ood} for the evaluation prompt). Similar to our prior experiments, we consider GPT-4o annotated judgments as the reference. From the results presented in Table \ref{tab:out_of_domain}, we find that the performance of the ChartJudge-2B model significantly deteriorates after fine-tuning, in comparison to its backbone Qwen2-VL-2B-Instruct model, with the 7B parameter Qwen2-VL performing the best. Therefore, while ChartJudge-2B improves performance in chart benchmarks, it lacks out-of-domain generalization.  

\begin{table}[t!]
\small

    \centering
    {
    \begin{tabular}{l|c}
        \toprule
        \textbf{Model} & \textbf{Judgment Accuracy} \\
        \midrule
        \rowcolor{models} ChartJudge-2B & 49.53 \\
        \rowcolor{open_models_below_4B!50} Qwen2-VL-2B-Instruct & 80.26 \\
        \rowcolor{open_models_7B_12B!50} Qwen2-VL-7B-Instruct & \textbf{90.64}  \\
   
        \bottomrule
    \end{tabular}
    }
    \caption{Results based on pair-wise comparison across 3K samples in the ChEBI-20-MM dataset for \textit{out-of-domain generalization}  evaluation for different models.}
    \label{tab:out_of_domain}
\end{table}

\subsubsection{Inference Cost and Latency Analysis} 

In this section, we do some cost and latency analyses. While our ChartJudge model could be deployed in a machine with just 8GB VRAM, other powerful baselines like Qwen-2-VL-7B-Instruct or LLaVA-Critic-7B require GPUs having at least 24GB VRAM. Therefore, ChartJudge-2B can significantly reduce the deployment cost when used as the LVLM judge to evaluate other LVLMs in real-world industrial settings. Moreover, to ensure efficiency in the CI/CD pipeline, it would also be useful to have a judge model having faster inference. In this case, ChartJudge-2B again demonstrates effectiveness. Based on our evaluation on a machine with 1 A30 GPU, we find that the throughput of ChartJudge-2B is approximately 110 tok/sec (8.1 ms/token latency). On the contrary, the throughputs for the 7B models (e.g., Qwen-2-VL-7B-Instruct, LLaVA-Critic-7B) are approximately 63 tok/sec (15.9 ms/token latency). Moreover, in terms of the inference cost, 7B models can be deployed in GCP\footnote{\url{https://cloud.google.com/compute/gpus-pricing}, last accessed: July 4, 2025} on a machine with 1 L4 GPU with 24GB VRAM, which costs $\$0.7$ to $\$0.9$ per hour; whereas 2B models can be deployed using T4 GPU that costs $\$0.3$ to $\$0.5$ per hour. This makes ChartJudge-2B twice as fast (while being \textbf{2x} cheaper) than the 7B parameter LVLM judges. We also provide further cost and latency benchmarking in Appendix \ref{cost_latency_benchmark}. 

\subsection{Deployment Considerations}
Despite its tiny size, ChartJudge-2B emerges as a pragmatic industrial judge, especially in pairwise scenarios. Since its pointwise error distance remains comparatively high, we recommend deploying it as an efficient pairwise judge that compares the existing deployed model against the newly developed model for pairwise comparison. Moreover, the model’s ability to perform evaluation based on multi-criteria prompting enables cost-efficient inference, which can significantly reduce the usage cost. 
While the model ChartJudge-2B loses its zero-shot ability when evaluated on an out-of-domain molecular dataset, the \textit{ChEBI-20-MM} benchmark, a brief fine-tune on synthetic judgments following our proposed domain adaptive transfer learning technique can be effective for adaptation across more tasks rather than being a deployment blocker.
\section{Conclusion and Future Work}
This paper proposes two cost-efficient strategies to improve the usability of tiny LVLMs as evaluation judges in chart comprehension tasks: \textbf{(i)} multi-criteria prompting, and \textbf{(ii)} fine-tuning via knowledge distillation from larger LVLMs to compact 2B-parameter models for better domain adaptation. Our experiments reveal that while various 7B LVLM judges suffer significant performance degradation under multi-criteria settings, our proposed ChartJudge-2B achieves 
superior performance with just 2B parameters 
 than many larger LVLMs in both single and multi-criteria settings. 
However, increased bias, with limited improvements in the pointwise evaluation imply the constraints of training on synthetic judgments. In the future, we plan to develop more flexible and modular prompting strategies that can be programmatically adjusted to different application requirements, allowing for dynamic evaluation across diverse chart-related tasks \cite{rahman2025text2vis,mahbub2025perils}, as well as other domains \cite{laskar-etal-2025-improving}. Furthermore, to address the increased bias observed in the fine-tuned model, we will conduct experiments to mitigate length bias. Finally, we will continue to study how to improve synthetic judgment data to 
further improve overall performance. 

\section*{Limitations}

While our proposed fine-tuned model may not always outperform the larger LVLMs that are used as the baseline, the main focus of this work is to demonstrate the effectiveness of knowledge distillation in the low-resource setting in the chart domain for LVLM-judges. Meanwhile, our proposed technique helps the model to outperform other models of similar size, alongside demonstrating robustness. 
Moreover, due to the rapid progress in LLM and LVLM development, there may have other tiny models that could be studied instead of Qwen2-VL-2B-Instruct model. However, we select Qwen2-VL due to the effectiveness of this model in prior research \cite{laskar2025judgingjudgeslargevisionlanguage} (see Appendix \ref{appendix:data_model_selection_criteria} for our explanation on model selection).  Since fine-tuning LVLMs require huge computing resources, we also focused to demonstrate the effectiveness of a 2B parameter model considering real-world scenarios. 
Moreover, for multi-criteria prompting, we only tested with factual correctness and informativeness since our focus was just to investigate how the LVLMs perform in the multi-criteria scenario in terms of the two commonly used qualitative evaluation metrics \cite{masry-etal-2023-unichart}. 
Nonetheless, future work may investigate the performance in other cases (e.g., relevance, conciseness) in the multi-criteria prompting approach.


\section*{Acknowledgments}
We thank all the anonymous reviewers of the EMNLP Industry Track for their excellent review comments. 
This research is supported by the Natural Sciences and Engineering Research Council (NSERC) of Canada, the York Research Chairs (YRC) program,  Canada Foundation for Innovation (CFI), Google's Gemini Academic Program for the API Credits, and Compute Canada for the computing resources. 

\section*{Ethical Statements}
The models used for experiments do not pose any ethical concerns since they are only used as the judge to evaluate other LVLM-generated responses. Additional compensation for human evaluation is not needed since it was conducted by the authors of this paper. Licensing requirements were maintained accordingly while using different artifacts (OpenAI, Gemini, Anthropic, HuggingFace).

\bibliography{custom}

\appendix

\section{Appendix}
\label{sec:appendix}

\subsection{Regarding Model and Dataset Selection}
\label{appendix:data_model_selection_criteria}
For the model and dataset selection, we follow the work of \cite{laskar2025judgingjudgeslargevisionlanguage}. However, we remove the models for our experiments that perform poorly in their work (e.g., larger models with at least 7B parameters that achieve below 50\% judgment performance). Based on this criterion, we selected the following models as the baseline: \textbf{(i)} \texttt{XGen-MM-Phi-3-3.8B} \cite{xue2024xgen} – \textit{a multimodal model (3.8B) developed by Salesforce}, 
\textbf{(ii)} \texttt{Phi-3.5-3.8B-Vision-Instruct}  \cite{abdin2024phi} – \textit{a smaller vision model from Microsoft}, \textbf{(iii)} \texttt{Qwen2-VL-2B}  - \textit{Alibaba’s Qwen \cite{wang2024qwen2} multimodal model with just 2B parameters}, \textbf{(iv)} \texttt{Qwen2-VL-7B} – \textit{The 7B version of the multimodal Qwen model}, \textbf{(v)} \texttt{InternLM-XComposer-7B} \cite{dong2024internlm} – \textit{a 7B vision model with composition abilities}, \textbf{(vi)} \texttt{LLaVA-v1.6-Mistral-7B} – \textit{A multimodal LVLM based on the LLaVA \cite{li2024llava} architecture that also utilizes a 7B Mistral \cite{jiang2023mistral7b} as the backbone}, and the \textbf{(vii)} \texttt{LLaVA-Critic-7B} – \textit{a specialized evaluator model based on LLaVA and Qwen}.

\subsection{Evaluation Process}
\label{definition_metrics}

In this section, we describe our evaluation process in detail. Note that we compute the results for different LVLMs across various evaluation metrics by comparing with GPT-4o and LLaVA-Critic-70B annotated judgments, and use the average score as reference to compare the performance of the LVLMs. Below, we illustrate the evaluation rubric and metrics. 

\paragraph{(i) Based on Evaluation Type:}\mbox{}\\

    - \textbf{Pairwise:} Judge selects the better of the two model-generated responses.
     
    - \textbf{Pointwise:} Judge rates a single response on a 1--5 Likert scale.

\paragraph{(ii) Based on Reference Type:}\mbox{}\\

    - \textbf{With Reference:} Ground-truth answer is provided with the chart and the response for judgment.
       
    - \textbf{Without Reference:} Judge relies only on the chart and response(s).

\paragraph{(iii) Based on Evaluation Criteria:}\mbox{}\\
  
    - \textbf{Factual Correctness:} Assesses factual accuracy of the response.

    - \textbf{Informativeness:} Measures the usefulness of the information in the response.

    - \textbf{Relevance:} Evaluates whether the response is relevant to the chart, and the query (in QA tasks).

    - \textbf{Multidimensional Evaluation:} Considers overall response quality by considering factual correctness, informativeness, and relevance.

\paragraph{(iv) Based on Evaluation Metrics:}\mbox{}\\
  
    - \textbf{Judgment Accuracy:} Check for exact match between the model-predicted judgment and the gold judgment (pairwise). 

    - \textbf{Error Distance:} Measures the average deviation from the reference rating (pointwise). 

    - \textbf{Bias Metric:} Checks if judgment changes with response order or judge mistakenly prefers the wrong answer due to bias towards lengthy responses (pairwise). 

    - \textbf{Instruction Following Evaluation Accuracy:} Measures the judge's ability to assess instruction following capabilities of other LVLMs in chart-related tasks. 

    - \textbf{Format Adherence Accuracy:} Evaluates whether the response generated by the judge follows the required output format (e.g., JSON). 


\subsection{Fine-Tuning Dataset Statistics}
\label{chart_type_statistics}
We demonstrate the statistics of our fine-tuning dataset based on chart type in Table \ref{tab:dataset_statistics_based_on_chart_type}.

\begin{table}[t!]
\centering
\small
\begin{tabular}{lcc}
\toprule
\textbf{Type} & \textbf{\# Single-Criterion} & \textbf{\# Multi-Criteria} \\ 
\midrule
Bar & 7737  & 1857  \\ 
Line & 1622 & 792 \\ 
Pie & 148 & 98 \\ 
Table & 138 & - \\ 
Area & 51 & 51 \\ 
Scatter & 29 & 29 \\ 
\midrule
\end{tabular}
\caption{{Fine-tuning dataset statistics for single-criterion and multi-criteria settings based on Chart Type. Here, `\#' denotes total number of samples.}}
\label{tab:dataset_statistics_based_on_chart_type}
\end{table}

\begin{table}[t!]
\centering
\small
\begin{tabular}{lcccc}
\toprule
& \multicolumn{2}{c}{\textbf{\# Tokens}} & \multicolumn{2}{c}{\textbf{Avg. Time}} \\
\cmidrule(lr){2-3} \cmidrule(lr){4-5} 
\textbf{Model Size }& \textbf{CPU} & \textbf{GPU}& \textbf{CPU} & \textbf{GPU} \\
\midrule
7B Parameters  & 0.06 & 1.5 & 15.9 & 0.63 \\
3.8B Parameters & 0.13 & 3.3 & 7.8  & 0.30 \\
2B Parameters  & 0.40 & 3.7 & 2.5  & 0.25 \\
\bottomrule
\end{tabular}
\caption{Inference Benchmarking of different-sized LVLMs. Here, `\# Tokens' denotes the average number of tokens per second, while `Avg. Time' denotes the average time (in seconds) to generate each token.}
\label{tab:benchmark_speed}
\end{table}

\subsection{Annotation of Complex Charts \& Queries}
\label{appendix_complex_human}

\definecolor{attachedColor}{HTML}{e0efff}
\definecolor{attachedColor2}{HTML}{f3f3f3}
\definecolor{attachedColor3}{HTML}{FFE5CC}
\definecolor{attachedColor4}{HTML}{FFCCCC}
\begin{tcolorbox}[
boxrule=0.25pt,   
  colback=attachedColor,    
  colframe=black,           
  colbacktitle=attachedColor4, 
  coltitle=black,           
  title={{Human Annotation Guidelines}},
  fonttitle=\bfseries,      
  fontupper=\small          
]

\textbf{Chart and Question Categorization:}
To evaluate model performance, you are required to annotate chart images and questions based on complexity:

\begin{itemize}
    \item 
\textbf{Simple Charts:} Contain two columns (representing dependent and independent variables).
\item \textbf{Complex Charts:} Contain more than two columns, depicting multiple variables and visual elements.
\end{itemize}

\begin{itemize}
    \item \textbf{Simple Questions:} Focus on single-step data extraction.
    \item \textbf{Complex Questions:} Require multi-step reasoning, comparisons, and logical inferences.
\end{itemize}

\end{tcolorbox}

\subsection{Question Generation Prompt}
\label{appendix_pew_question}

\definecolor{attachedColor}{HTML}{e0efff}
\definecolor{attachedColor2}{HTML}{f3f3f3}
\definecolor{attachedColor3}{HTML}{FFE5CC}
\definecolor{attachedColor4}{HTML}{FFCCCC}
\begin{tcolorbox}[
boxrule=0.25pt,   
  colback=attachedColor2,    
  colframe=black,           
  colbacktitle=attachedColor3, 
  coltitle=black,           
  title={{Chart-to-Text Dataset (Pew Split)}},
  fonttitle=\bfseries,      
  fontupper=\small          
]

Generate a concise question related to the summary of the given chart. \\

        [Model Generated Summary] \\

        [\texttt{Chart Image}]

\end{tcolorbox}

\subsection{Benchmarking Inference Cost and Latency}
\label{cost_latency_benchmark}

We also benchmarked the inference speed of models of varying sizes on an AMD machine with a 64-core processor, testing both GPU (1 A100) and CPU-only modes. As shown in Table \ref{tab:benchmark_speed}, smaller models exhibit significantly faster inference, especially in resource-constrained, CPU-only environments. Our 2B model achieves an average time per token of 2.5 seconds on a CPU, whereas a 7B model takes 15.9 seconds. This highlights the real-world value of using a fine-tuned tiny model for efficient, low-latency evaluation.

\subsection{Multi-Criteria Prompts for LVLM Judge}
\label{appendix_prompt}


\definecolor{attachedColor}{HTML}{e0efff}
\definecolor{attachedColor2}{HTML}{f3f3f3}
\definecolor{attachedColor3}{HTML}{FFE5CC}
\definecolor{attachedColor4}{HTML}{FFCCCC}
\begin{tcolorbox}[
boxrule=0.25pt,   
  colback=attachedColor2,    
  colframe=black,           
  colbacktitle=attachedColor3, 
  coltitle=black,           
  title={{OpenCQA Pointwise (With Reference)}},
  fonttitle=\bfseries,      
  fontupper=\small          
]

Suppose, you are a human evaluator. 
  You are required to rate the \textbf{Informativeness} and \textbf{Factual Correctness} of the answer generated by a model in comparison to the gold reference answer for a given question in the open-ended chart question answering task. \\

        Provide your response in an Array of JSON with the following keys: (i) Score, (ii) Explanation, (iii) Type. \\

        Here, the `Score' should be between 1 to 5 (inclusive), with the higher score indicating better. Moreover, the value for the `Explanation' key will contain the reasoning behind your scoring, while the value for the `Type' key will contain either `informativeness' or `factual correctness', depending on the type for which you assess the answer.  \\


        In the following, you are first given the question, followed by the gold reference answer. 
        Afterward, you are given the model-generated answer. You are also provided with the chart image as the context for the chart question-answering task. \\

        [Question] \\

        [Gold Reference Answer] \\

        [Model Generated Answer] \\

        [\texttt{Chart Image}]

\end{tcolorbox}

\begin{tcolorbox}[
boxrule=0.25pt,   
  colback=attachedColor2,    
  colframe=black,           
  colbacktitle=attachedColor3, 
  coltitle=black,           
  title={{OpenCQA Pairwise (Without Reference)}},
  fonttitle=\bfseries,      
  fontupper=\small          
]

Suppose, you are a human evaluator. 
         You are given the answers generated by two different models for a given question in the open-ended chart question answering task. Now, your task is to determine which model is better in terms of \textbf{Informativeness} and \textbf{Factual Correctness}. \\

        Provide your response in an Array of JSON with the following keys: (i) Model, (ii) Explanation, (iii) Type.
\\

        Here, the output value for the `Model' key is the respective model that is better, could be either `Model A' or `Model B', or `Tie' if both models are equally good. Moreover, the value for the "Explanation" key will contain the reason behind your preference, while the value for the `Type' key will contain either `informativeness' or `factual correctness', depending on the type for which you assess the answer.  
\\


        In the following, you are first given the question. 
        Afterward, you are given the model-generated answers. You are also provided with the chart image as the context for the chart question-answering task. \\

        [Question] \\

        [Model A Generated Answer] \\

        [Model B Generated Answer] \\

        [\texttt{Chart Image}]

\end{tcolorbox}

\subsection{Sample Response in Multi-Criteria Prompting}
\label{appendix_response}


\begin{tcolorbox}[
boxrule=0.25pt,   
  colback=attachedColor2,    
  colframe=black,           
  colbacktitle=attachedColor3, 
  coltitle=black,           
  title={{Response in OpenCQA (Pairwise)}},
  fonttitle=\bfseries,      
  fontupper=\small          
]

[\\
    \{"Model": "Model B", \\
        "Explanation": "Model B provides a thorough breakdown of the Mexican-American population growth, with the numbers accurately matching the chart.",\\
        "Type": "Informativeness"\\
    \},\\
    \{\\
        "Model": "Model A",\\
        "Explanation"": "Model A is factually correct as it accurately reports the figures for both new immigrants and births during the decade 2000-2010.",\\
        "Type": "Factual Correctness"\\
    \}\\
]
\end{tcolorbox}

\subsection{LLM-Judge Prompt for Out-of-Domain Generalization Evaluation}
\label{appendix_ood}

\definecolor{attachedColor}{HTML}{e0efff}
\definecolor{attachedColor2}{HTML}{f3f3f3}
\definecolor{attachedColor3}{HTML}{FFE5CC}
\definecolor{attachedColor4}{HTML}{FFCCCC}
\begin{tcolorbox}[
boxrule=0.25pt,   
  colback=attachedColor2,    
  colframe=black,           
  colbacktitle=attachedColor3, 
  coltitle=black,           
  title={{Pairwise Evaluation on ChEBI-20-MM for Molecular Image Captioning)}},
  fonttitle=\bfseries,      
  fontupper=\small          
]

You are given the summaries generated by two different models in the molecular image captioning task. \

Now, your task is to determine which model is better by considering the relevancy, conciseness, informativeness, and factual correctness of the model generated caption. \\

Ideally, a better quality caption should be concise, factually correct, and provide high level informative insights of the molecular image. \\

Please provide your response in JSON format with the following keys: (i) Model, (ii) Explanation. \\

Here, the output value for the 'Model' key is the respective model that is better, could be either 'Model A' or 'Model B', or 'Tie' if both models are equally good. \\

The value for the ""Explanation"" key will contain the reason behind your preference. \\

You should only provide the response in the required JSON format without any additional text such that I can correctly parse the result from your JSON formatted response. \\

In the following, you are given the model generated captions. You are also provided with the molecular image which was used as the context for the molecular image captioning task.\\

[Model A Generated Caption] \\

[Model B Generated Caption] \\

[Chart Image]

\end{tcolorbox}


\subsection{Error Examples}

Some error examples in multi-criteria settings for different LVLMs are shown in Table \ref{tab:error_examples}.

\label{error_examples}
\begin{table*}[t]
\centering
\scriptsize
\renewcommand{\arraystretch}{1.2}
\begin{tabular}{p{0.2\textwidth}p{0.38\textwidth}p{0.3\textwidth}}
\toprule
\textbf{Model} & \textbf{Response} & \textbf{Remarks} \\
\midrule
\textbf{Qwen2-VL-2B-Instruct} & \{ "Model A": \{ "Explanation": "The model generated answer is informative and factual, as it provides a detailed breakdown of the data from the chart, including the percentage of adults in each G20 country who see climate change as a major threat. It also provides insights into the varying perspectives on the issue across the G20 group.", "Type": "informativeness" \}, "Model B": \{ "Explanation": "The model generated answer is informative, as it provides a summary of the data from the chart, including the percentage of adults in each G20 country who see climate change as a major threat. However, it does not provide any insights into the varying perspectives on the issue or the factors shaping these views. It also does not provide any additional information beyond what is already in the chart.", "Type": \textcolor{red}{"informativeness"} \} \} & Both types are predicted as informativeness, with no factual correctness. Same error also happens in the 7B model. \\
\midrule
\textbf{InternLM-Xcomposer2d5-7B} & [ \{ "Model": "Model A", "Explanation": "Model A's answer is informative and provides a comprehensive overview of the data, including both the highest and lowest percentages of concern about climate change among the G20 countries. It also includes an analysis of the potential factors behind these differences and suggests further areas for exploration. This aligns well with the chart's information and demonstrates a high level of understanding of the data presented. Therefore, Model A is more informative than Model B, which focuses solely on the factual data without additional insights or analysis. Factual correctness is not an issue in this context as both models accurately reflect the data from the chart. \textcolor{red}{Type: informativeness.}" \}, \{ "Model": "Model B", "Explanation": "Model B's answer is factually correct but lacks the depth and insight provided by Model A. It simply reiterates the data from the chart without offering any additional analysis or interpretation. While it is accurate, it does not provide the same level of value to the user seeking deeper understanding of the data. Factual correctness is maintained, but the lack of additional insights means it falls short in terms of informativeness compared to Model A. \textcolor{red}{Type: factual correctness.}" \} ] & Double quotes missing for the Type, leading to JSON parsing error. Similar error also happens in the LLaVA-Critic-70B model.  \\
\midrule
\textbf{LLaVA-Next-v1.6-Mistral-7B} & \textcolor{red}{\{ "Model": "Model A", "Type": "informativeness", "Explanation": "Model A provides a more comprehensive and accurate explanation of the chart, highlighting the significant difference in the proportion of broadband and dial-up users who get their news on the internet. It also explains the reasons behind the observed difference, which is not present in Model B's answer." \}} \newline \textcolor{red}{\{ "Model": "Model A", "Type": "factual correctness", "Explanation": "Model A's answer is factually correct, as it accurately reflects the data presented in the chart. Model B's answer is incorrect, as it states that dial-up users are more likely to get their news on the internet than broadband users, which contradicts the facts given in the chart." \}}  & Wrong format, responded as JSON lines instead of array of JSON.\\
\midrule
\textbf{LLaVA-Critic-7B} & \textcolor{red}{['```json [ \{ "Model": "Model A", "Explanation": "Model A correctly identifies that broadband users (41\%) have a significantly higher proportion of internet news users compared to dial-up users (23\%). This explanation aligns with the data in the chart, providing a clear and accurate comparison between the two types of internet users.", "Type": "informativeness" \}, \{ "Model": "Model B", "Explanation": "Model B incorrectly states that 41\% of dial-up users get their news on the internet, which contradicts the chart data showing 23\%. Additionally, it incorrectly concludes that dial-up users are more likely to get their news on the internet than broadband users, which is not supported by the data. This answer is factually incorrect.", "Type": "factual correctness" \} ] ``` Model A provides an accurate and informative comparison between broadband and dial-up users based on the chart data, while Model B contains factual inaccuracies and misinterprets the data. Therefore, Model A is better in terms of both informativeness and factual correctness.']} & Wrong format, could not follow the JSON format instruction properly. Similar error also happens in the 70B model. \\
\bottomrule
\end{tabular}
\caption{Examples of errors in Multi-Criteria settings by different LVLMs.}
\label{tab:error_examples}
\end{table*}

\end{document}